\documentclass[sigconf]{acmart}
\usepackage{float}
\usepackage{tabularx}
\usepackage{tabularx}
\usepackage{booktabs}
\usepackage{enumitem}
\usepackage{array}
\newcolumntype{Y}{>{\raggedright\arraybackslash}X}
\usepackage{makecell}
\usepackage{hyperref}
\usepackage{url}
\usepackage[table]{xcolor} 
\usepackage{colortbl}
\usepackage{multirow}   
\usepackage{booktabs}    
\newtheorem{myDef}{Definition}
\usepackage[most,skins,theorems]{tcolorbox}
\tcbset{
  aibox/.style={
    width=\linewidth,  
    top=7pt,
    bottom=2pt,
    colback=blue!6!white,
    colframe=black,
    colbacktitle=black,
    enhanced,
    center,
    attach boxed title to top left={yshift=-0.1in,xshift=0.15in},
    boxed title style={boxrule=0pt,colframe=white,},
  }
}
\newtcolorbox{AIbox}[2][]{aibox,title=#2,#1}
\usepackage{graphicx}
\usepackage{booktabs}    
\usepackage{caption}

\usepackage{graphicx} 
\usepackage{arydshln}

\newtheorem{pro}{Problem}

\newcommand{\thinkstart}[1]
{\textcolor{gray}{\texttt{<think>}}}

\newcommand{\thinkend}[1]
{\textcolor{gray}{\texttt{</think>}}}

\newcommand{\tollcallstart}[1]
{\textcolor{gray}{\texttt{<tool\_call>}}}

\newcommand{\tollcallend}[1]
{\textcolor{gray}{\texttt{</tool\_call>}}}

\newcommand{\tollresponsestart}[1]
{\textcolor{gray}{\texttt{<tool\_response>}}}

\newcommand{\tollresponseend}[1]
{\textcolor{gray}{\texttt{</tool\_response>}}}

\newcommand{\answerstart}[1]
{\textcolor{gray}{\texttt{<answer>}}}

\newcommand{\answerend}[1]
{\textcolor{gray}{\texttt{</answer>}}}

\AtBeginDocument{%
  }

\copyrightyear{2026}
\acmYear{2026}
\setcopyright{cc}
\setcctype{by}
\acmConference[KDD 2026] {Proceedings of the 32nd ACM SIGKDD Conference on Knowledge Discovery and Data Mining V.2}{August 9--13, 2026}{Jeju Island, Republic of Korea.}
\acmBooktitle{Proceedings of the 32nd ACM SIGKDD Conference on Knowledge Discovery and Data Mining V.2 (KDD 2026), August 9--13, 2026, Jeju Island, Republic of Korea}
\acmISBN{979-8-4007-2259-2/2026/08}
\acmDOI{10.1145/3770855.3818359}

\begin{document}

\title{DeepTravel: An End-to-End Agentic Reinforcement Learning Framework for Autonomous Travel Planning Agents}

\author{Yansong Ning}
\affiliation{%
  \institution{The Hong Kong University of Science and Technology (Guangzhou)}
  \city{Guangzhou}
  \state{Guangdong}
  \country{China}
}
\email{yning092@connect.hkust-gz.edu.cn}

\author{Rui Liu}
\affiliation{%
  \institution{Didichuxing Co. Ltd}
  \city{Beijing}
  \country{China}
}
\email{invincibleliu@didiglobal.com}

\author{Jun Wang}
\affiliation{%
  \institution{Didichuxing Co. Ltd}
  \city{Beijing}
  \country{China}
}
\email{wangjun@didiglobal.com}

\author{Kai Chen}
\affiliation{%
  \institution{Didichuxing Co. Ltd}
  \city{Beijing}
  \country{China}
}
\email{kevinchenkai@didiglobal.com}

\author{Wei Li}
\affiliation{%
  \institution{Didichuxing Co. Ltd}
  \city{Beijing}
  \country{China}
}
\email{peterliwei@didiglobal.com}

\author{Jun Fang}
\affiliation{%
  \institution{Didichuxing Co. Ltd}
  \city{Beijing}
  \country{China}
}
\email{fangjun@didiglobal.com}

\author{Kan Zheng}
\affiliation{%
  \institution{Didichuxing Co. Ltd}
  \city{Beijing}
  \country{China}
}
\email{zhengkan@didiglobal.com}

\author{Naiqiang Tan}
\affiliation{%
  \institution{Didichuxing Co. Ltd}
  \city{Beijing}
  \country{China}
}
\email{tannaiqiang@didiglobal.com}

\author{Hao Liu}
\authornote{Corresponding author.}
\affiliation{%
  \institution{The Hong Kong University of Science and Technology (Guangzhou)}
  \city{Guangzhou}
  \state{Guangdong}
  \country{China}
}
\email{liuh@hkust-gz.edu.cn}


\renewcommand{\shortauthors}{Yansong Ning et al.}


\begin{abstract}
Travel planning (TP) agent has recently worked as an emerging building block to interact with external tools/resources for travel itinerary generation, ensuring an enjoyable user experience.
Despite its benefits, existing studies rely on hand-craft prompt and fixed agent workflow, hindering more flexible and autonomous TP agents.
This paper proposes \textbf{DeepTravel}, an end-to-end agentic reinforcement learning framework for building an autonomous travel planning agent, capable of autonomously planning, executing tools, and reflecting on tool responses to explore, verify, and refine intermediate actions in multi-step reasoning.
To achieve this, we first construct a robust travel sandbox by caching transportation, accommodation and POI data, facilitating TP agent training without being constrained by real‑world APIs limitations (e.g., inconsistent outputs).
Moreover, we develop a hierarchical reward modeling system, where a trajectory‑level verifier first checks spatiotemporal feasibility and filters unsatisfied travel itinerary, and then the turn‑level verifier further validate itinerary's detail consistency with tool responses, enabling efficient and precise reward service.
Finally, we propose the reply-augmented reinforcement learning method that enables TP agent to periodically replay from a failure experience buffer, emerging notable agentic capacity.
We deploy the trained TP agent in the DiDi Enterprise Solutions application. 
A three-month online test shows that it achieves 82\% accuracy in travel itinerary generation.
Comprehensive offline evaluations further demonstrate that DeepTravel enables small-sized LLMs (e.g., Qwen3-32B) to significantly outperform frontier LLMs (e.g., OpenAI o1/o3 and DeepSeek-R1) and existing TP agent frameworks.
\end{abstract}




\maketitle



%

\newcommand{\fix}{\marginpar{FIX}}
\newcommand{\new}{\marginpar{NEW}}


\section{Introduction}
Travel planning (TP) aims to create a feasible itinerary \cite{Nguyen2023TheIO} that aligns with user preference by integrating multiple resources, such as accommodations, transportation, and Points-of-Interest (POI).
Recently, with the advances in natural language processing, large language models (LLMs) are widely used to build TP agents \cite{chen2024travelagent}, capable of invoking external tools/resources \cite{gou2023tora} to generate travel itineraries, offering a seamless and enjoyable experience in human mobility \cite{tang2024itinera}.
As a result, the LLM-powered TP agent has gradually emerged as a popular tool for modern citizens.

\begin{figure}

    \centering
    \includegraphics[width=1\linewidth]{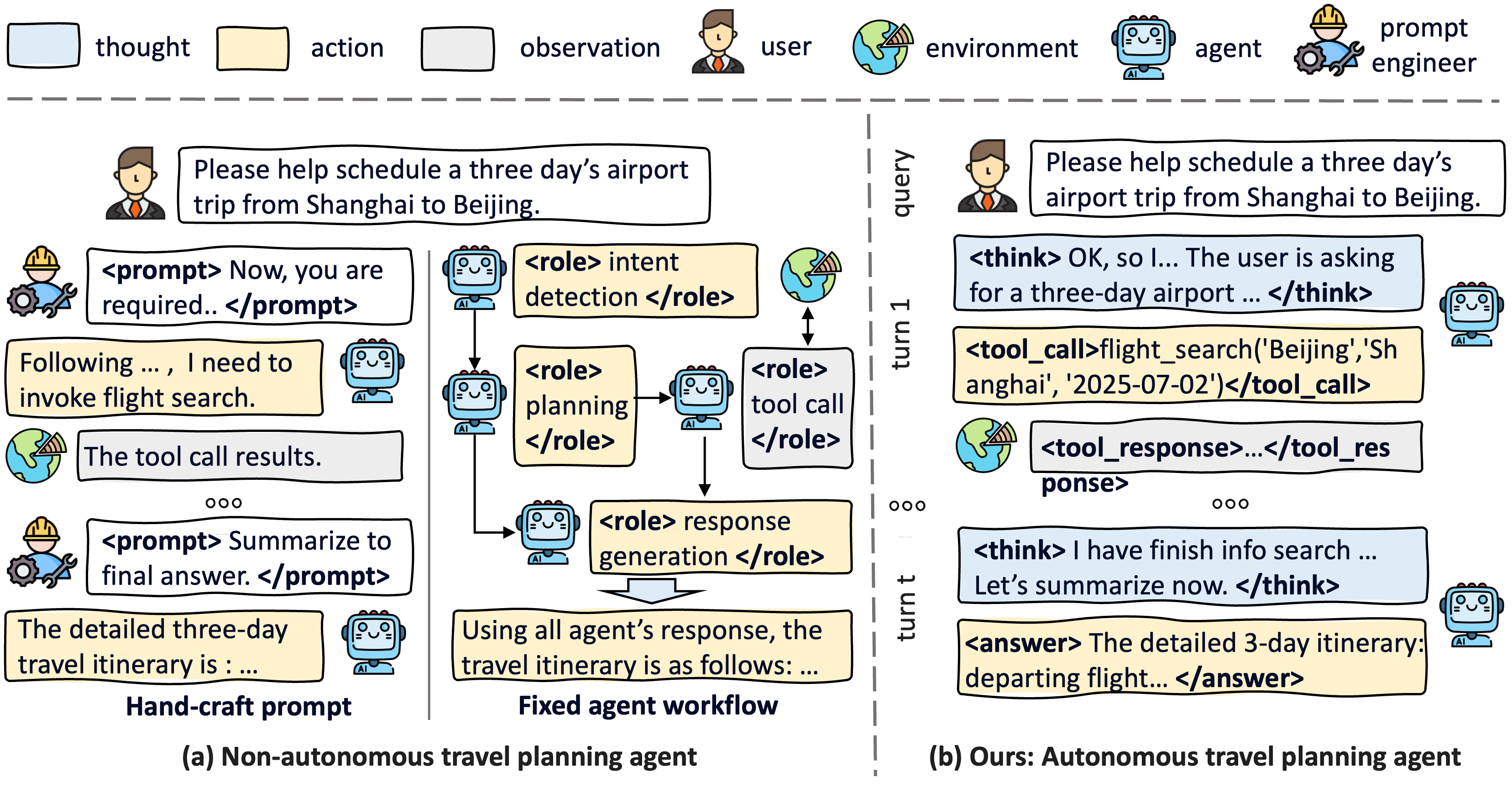}
    \caption{Comparison between existing studies and our autonomous travel planning agent paradigm.}
    \label{fig1_intro}
\end{figure}
In recent literature, many efforts have been devoted to construct travel planning agent.
As illustrated in Figure \ref{fig1_intro}(a), most existing approaches primarily rely on carefully designed prompts.
For example, TravelPlanner \cite{xie2024travelplanner} and TripTailor \cite{wang2025triptailor} employ task-specific prompts to guide LLMs for tool invocation and itinerary generation.
More recently, researchers have begun to integrate these prompt-engineering strategies into fixed agent workflow.
For instance, PTS \cite{shao2025personal} and RETAIL \cite{deng2025retail} propose well-structured agent pipelines that enhance user intention understanding, enable effective tool interactions, and support accurate travel itinerary generation.
However, these methods are \emph{labor-intensive} and face challenges in \emph{adapting to new user query or recovering from tool call failures}, limits more flexible and autonomous TP agents.

Agentic reinforcement learning (RL) \cite{singh2025agentic} has recently emerged and shown possibilities for building autonomous AI agent \cite{jaech2024openai} by enabling agent to interact with tools in a dedicated environment and refine its reasoning based on the feedback it receives.
For example, ReTool \cite{feng2025retool}, Kimi-Researcher \cite{kimiResearcher} and WebSailor \cite{li2025websailor} utilize an end-to-end agentic RL training to build the autonomous agent for math, deep research and web domain, respectively.
These studies motivate us to propose a tailored agentic reinforcement learning framework for autonomous TP agent construction, addressing the limitation in existing travel planning studies.

However, building an autonomous TP agent shown in Figure \ref{fig1_intro}(b), that can progressively tackle complex TP tasks by autonomously interleaving tool calls and tool responses within the multi-turn interaction, is a non-trivial problem due to two key factors.
\emph{(1) Dynamic Travel Environment.}
TP agents operate in a highly dynamic environment where information—such as hotel availability, pricing, and transportation options—changes continuously in real time.
As a result, identical queries may produce inconsistent outputs over time due to updates in accommodations, transportation, and POI data sources.
Training TP agents in such a constantly evolving real-world environment remains a significant challenge.
\emph{(2) Open-Ended Travel Task.}
Unlike existing reinforcement learning with verified rewards (RLVR) paradigm \cite{guo2025deepseek} on math or web domain, travel planning is an inherently open-ended task without explicit ground truth. 
For example, the generated travel itinerary may vary depending on personalized user preference and budget, making the outcome difficult to verify. 
Therefore, how to construct reliable and scalable reward signals is challenging.

To address the aforementioned challenges, we propose \textbf{DeepTravel}, an end-to-end agentic RL training framework for autonomous travel planning agent construction.
Specifically, we first construct a \emph{Robust Travel SandBox} by caching transportation, accommodation, and POI data from multiple real-world APIs across different timestamps, thereby simulating dynamic tool interactions. 
Within this sandbox, the TP agent can perform large-scale repeated trial‑and‑error learning while overcoming QPS limits and output inconsistencies.
Moreover, we propose a \emph{Hierarchical Reward Modeling} system, where a trajectory‑level verifier ensures the spatiotemporal feasibility of generated itineraries and a turn‑level verifier enforces fine‑grained consistency with tool response, thereby yielding more efficient and reliable reward signals for training.
Finally, we propose a \emph{Reply-Augmented Reinforcement Learning} method to incentivize agentic reasoning capacity through sequential cold-start and RL process. 
Based on periodically replay from a failures experience buffer, the TP agent can learn and refine its previous reasoning actions, gradually emerging agentic travel planning capacity.

We deploy constructed TP agent in DiDi Enterprise Solutions application, and conduct evaluation on real-world production environment.
Three months' online experiments demonstarte that DeepTravel-32B achieves 82\% accuracy in travel itinerary generation task.
We further conduct offline experiments using synthetic data across varying task complexity.
The experimental results demonstrate that DeepTravel enables small-size LLM backbones (e.g., Qwen3-32B) to outperform current state-of-the-art reasoning LLMs (e.g., OpenAI-o1/o3 and DeepSeek-R1) and prevailing RL algorithms (e.g., GRPO and DAPO).
These results establish DeepTravel as an promising framework to build autonomous TP agent.

Our contributions are summarized as follows: 
(1) We establish and deploy the first autonomous travel planning agent, offering new paradigm to advance existing TP studies.
(2) We propose the first end-to-end agentic RL framework tailored to travel domain, which allows the training of TP agent under a roboust sandbox environment, reliable reward service and periodical experience replay strategy.
(3) Extensive online and offline test validate the effectiveness of proposed framework and uncover its exceptional performance across travel planning tasks.

\begin{table*}[]
\caption{The specifications of unified toolkit in travel sandbox.}
\label{toolkits}
\centering
\begin{tabular}{cccc}  
\toprule  
Type                            & Tool name      & Tool call format & Tool response description \\
\midrule
\multirow{3}{*}{Transportation} & flight search   & flight\_search(depart\_city\_name, arrival\_city\_name, depart\_date) & feasible flight options \\
                                & train search   & train\_search(depart\_city\_name, arrival\_city\_name, depart\_date) & feasible train options \\
                                & route planning & route\_planning(origin\_name, destination\_name, city\_name) & route, distance and time \\
\midrule
Accommodation                   & hotel search   & hotel\_search(city\_name, hotel\_name, checkin\_date, checkout\_date) & available hotel condidate \\
\midrule
\multirow{2}{*}{Attraction}     & POI search     & poi\_search(query, city\_name) & detailed address of POI \\
                                & web search     & web\_search(query) & web page related to the query  \\
\bottomrule  
\end{tabular}
\end{table*}

\section{Preliminary}
We begin with the definition of user query and travel itinerary, then define the problem we aim to address.

\begin{myDef} \textbf{User Query.}
The user query $q$ is expressed in natural language, which indicates user's spatiotemporal travel intention and personalized preference.
For example, a query ``Please help schedule a three day's airport trip from Shanghai to Beijing'' represents that the user wants to travel to Beijing by air and stay there for a duration of three days.
\end{myDef}

\begin{myDef} \textbf{Travel Itinerary.}
A travel itinerary $I$ is defined as a structured plan including accommodation, transportation, and detailed daily plan that integrates such as travel activity suggestions, exploration strategies for Points of Interest (POIs), and etc.
\end{myDef}

Note that the POIs usually corresponds to popular tourist spots (e.g., \emph{National Palace Museum}, \emph{The Great Wall}, etc.).
Now we formulate our problem:

\begin{pro} \textbf{Agentic Travel Planning.}
Given a query $q$, the travel planning agent generates travel itinerary $I$ to satisfy the trip requirements through automatically planing, executing tools, and reflecting on tool responses to explore, verify and refine intermediate actions in multi-turn interaction process:
\begin{equation}
\begin{aligned}
\left \{  \tau_{t}, a_{t}\right \}=\pi_{\theta} \left(q, \left \{ \tau_{1}, a_{1}, o_{1}, \tau_{2}, a_{2}, o_{2}, \ldots, \tau_{t-1}, a_{t-1}, o_{t-1}\right \}\right),
\end{aligned}
\end{equation}
where $\pi_{\theta}$ is the policy of travel planning agent, and $\tau_{t-1}, a_{t-1}, o_{t-1}$ represent agent's thought, action, and observation from the environment in the $t-1$ turn, respectively.
The generated travel itinerary $I$ is involved in the agent's action $a_{t}$ in the last turn.
\end{pro}

An agentic travel planning example is illustrated in Figure \ref{fig1_intro}(b), where the TP agent autonomously think (i.e., thought wrapped with \thinkstart{} and \thinkend{}) before using external tools (i.e., action enclosed with \tollcallstart{} and \tollcallend{}) and reflect on tool response (i.e., observation wrapped with \tollresponsestart{} and \tollresponseend{}) to explore, verify and refine intermediate step in multi-turn interaction process for generating travel itinerary.

\section{DeepTravel}
With these notations, we propose DeepTravel, an end-to-end agentic reinforcement learning pipeline for TP agent construction. 

\subsection{Framework Overview}
\begin{figure*}
    \centering
    \includegraphics[width=1\linewidth]{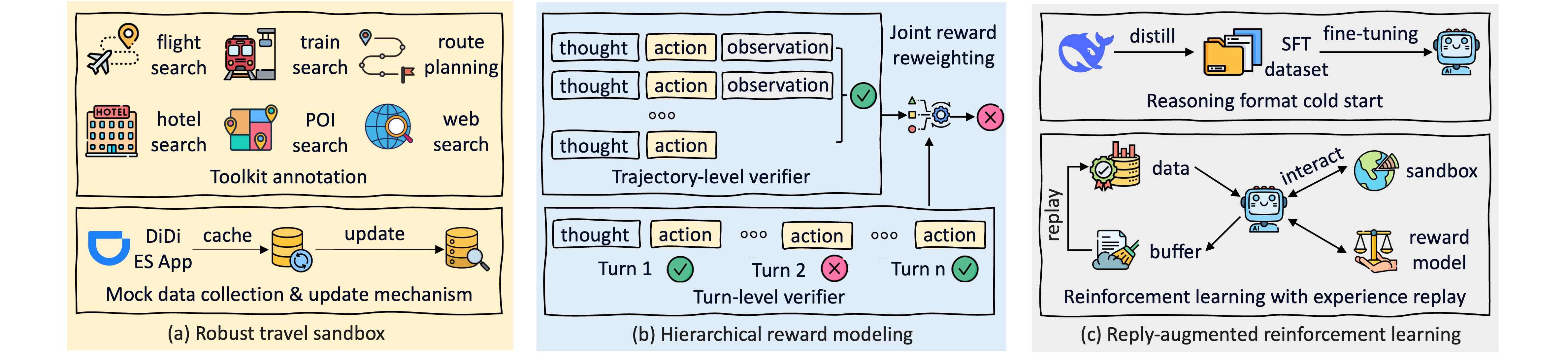}
    \caption{An overview of DeepTravel.}
    \label{fig_method}
            \vspace{-5pt}
\end{figure*}

Figure \ref{fig_method} illustrates the overall pipeline of DeepTravel.
\emph{(1) Robust Travel SandBox} involves toolkit annotation, mock data collection and update mechanism, thereby enabling simulated real-world tool interactions.
\emph{(2) Hierarchical Reward Modeling} proposes both of trajectory-level and turn-level verifier, which jointly provides reliable and efficient reward signal.
\emph{(3) Reply-Augmented Reinforcement Learning} begins with supervised fine-tuning to cold-start the agent in an agentic travel planning format, followed by replay-augmented reinforcement learning to incentivize the LLM’s agentic capabilities.

\subsection{Robust Travel SandBox}
The sandbox \cite{lin2023agentsims} is served as a stable environment for TP agent to interact with tools, simulating real-world interaction while overcoming practical output inconsistency and API limits. 
We first introduce how real-world tools are abstracted into a unified execution interface, and then describe the sandbox data collection and update mechanism built on top of this interface.

\subsubsection{Unified Toolkit Interface Construction}
We unify six categories of real-world tools into a executable interface for the TP agent.
Table \ref{toolkits}  summarizes the unified parameter schema and invocation format that standardizes heterogeneous tool APIs into a consistent execution interface.

\textbf{Flight Search.} 
Flight search offers information about air transportation, a fundamental aspect of travel planning \cite{shao2024chinatravel}.
Instead of directly exposing the original DiDi ES API, we re-structure its interface into the unified tool schema defined by our sandbox, ensuring invocation for the TP agent.
Each call follows the standardized format with departure city, arrival city, and departure date.
For example, \emph{``flight\_search(`Beijing', `Shanghai', '2025-07-02')''} queries for flight options from \emph{`Beijing'} to \emph{`Shanghai'} on July 2, 2025.

\textbf{Train Search.}
Train search represents rail transportation services in real-world travel planning.
Similar to flight search, train search provides essential rail transportation information. 
We follow the DiDi ES interface format to build this tool. 
As shown in Table \ref{toolkits}, each query includes a departure city, an arrival city, and a departure date.
For example, \emph{``train\_search(`Beijing', `Shanghai', '2025-07-02')''} queries for train options from \emph{`Beijing'} to \emph{`Shanghai'} on July 2, 2025.

\textbf{Route Planning.}
Route planning provides spatial routing and travel cost estimation based on map services, which is crucial for optimizing travel time and cost \cite{fang2024travellm}. 
We use the DiDi Map routing API to build this unified tool interface, transforming map-specific request into the standardized parameter schema \cite{ning2025dima}.
Each invocation includes an origin name, a destination name, and a city name, allowing the TP agent to obtain distance and time estimation results.
For instance, \emph{``route\_planning(`National Palace Museum', `The Great Wall', 'Beijing')''} plans the route and calculates the distance/time from the \emph{`National Palace Museum'} to \emph{`The Great Wall'} in Beijing.

\textbf{Hotel Search.}
Hotel search enables the TP agent to find suitable accommodations based on user preferences \cite{yang2025plan}. 
Similarly, we also leverage the DiDi ES interface to construct this tool. 
Each tool call includes a city name, a hotel name, a check-in date, and a check-out date. 
For example, \emph{``hotel\_search(`Beijing', `Atour', '2025-07-02', '2025-07-05')''} searches for available rooms at the \emph{`Atour'} hotel in \emph{`Beijing'} from July 2 to July 5, 2025.

\textbf{POI Search.}
POI search provides urban contextual semantics, which has been widely adopted in travel planning \cite{xie2024travelplanner}.
Based on DiDi Map, we build a unified tool schema, converting its original query structure into a standardized invocation format compatible with the sandbox execution environment.
Each tool call contains a query and a city name.
For example, \emph{``poi\_search(`The Great Wall', `Beijing')''} helps obtain the address of \emph{`The Great Wall'}.

\textbf{Web Search.}
Web search provides access to open-domain travel knowledge \cite{ning2024urbankgent} or plan, serving as a valuable resource for generating itineraries \cite{ni2025tp}. 
Instead of exposing raw web search APIs, we encapsulate the Bocha AI service into the unified tool interface.
The tool call parameter is any query related to travel planning (e.g., \emph{``web\_search(`Introduction to Beijing')''}).

\subsubsection{Mock Data Caching}
To simulate the dynamic nature of real-world tool interactions, where prices and availability of hotels, trains, and flights fluctuate over time, we propose a data caching mechanism. 
This mechanism addresses the challenge of real-world API QPS limits and inconsistent information retrieval during each search, which can hinder the agent's ability to learn from previously failed cases. 
Upon receiving a novel query, we will retrieve fresh data from the external API and persists it into the local database.

\subsubsection{Mock Data Update}
We employ an on-demand caching strategy with a 24-hour refresh cycle to maintain data relevance while ensuring training stability.
During RL training, this mechanism is vital for ``experience replay'', allowing the TP agent to re-access earlier tool response under identical conditions.
It also enables TP agent to repeatedly learn from unsuccessful cases through exploring, and refining its intermediate reasoning trajectories \cite{shang2025rstar2}.

\subsection{Hierarchical Reward Modeling}
Then, we present our reward modeling system.
This system comprises a trajectory-level verifier and a turn-level verifier, designed to provide efficient and reliable reward signals for agent training.

\subsubsection{Trajectory-Level Verifier.}
This verifier assesses the overall spatiotemporal feasibility of the generated travel itinerary. 
Given a complete reasoning trajectory $\left \{ \tau_{1}, a_{1}, o_{1}, \tau_{2}, a_{2}, o_{2}, \ldots, \tau_{t}, a_{t}\right \}$, the trajectory-level verifier checks whether the final travel itinerary $a_{t}$ adheres to essential spatiotemporal constraints \cite{chaudhuri2025tripcraft}. 
These constraints include such as logical sequence of events, geographic plausibility, and satisfaction of user requirements (e.g., visiting specific POIs within a given timeframe). 
This coarse-grained evaluation efficiently filters out invalid itineraries, ensuring that only potentially valid plans proceed to the next level of verification.


\subsubsection{Turn-Level Verifier.}
Upon successful verification by the above trajectory-level verifier, the turn-level verifier performs a more granular evaluation. 
This verifier examines the consistency between the agent's final travel itinerary $a_{t}$ and the information obtained by the external tools at each turn \cite{zeng2025reinforcing}. 
Specifically, for each turn $i$ from $1$ to $t-1$, the verifier assesses whether $a_{t}$ accurately reflect the tool response $o_{i}$.
By systematically verifying each turn, it helps identify factual hallucination/mistakes of LLM-generated travel itinerary.


\subsubsection{Joint Reward Reweighting.}
Finally, the results of two verifier will be combined using a joint reward reweighting strategy to provide reward signal. 
Specifically, if the trajectory-level verifier detects a violation, the final reward $r$ is immediately set to 0, saving computational resources. 
If the trajectory passes the trajectory-level verification, the turn-level verifier assesses each turn.
The final reward $r$ is set to 1 only if every turn passes verification, indicating a fully consistent and valid travel itinerary. 

This hierarchical structure ensures both the efficiency and reliability of the reward modeling system.
In practical implementation, we build many travel-oriented rubrics \cite{huang2025reinforcement} for trajectory-level and turn-level verifier, respectively.
Based on human generated rubrics, we prompt DeepSeek-R1 based verifier to provide reward modeling service.
Details could be found in Appendix \ref{appendix_prompt}.

\subsection{Replay-Augmented Reinforcement Learning}
This section details relay-augmented reinforcement learning, which is a two-stage process.
We first employ supervised fine-tuning (SFT) to initialize reasoning format of TP agent.
Then, we leverage RL to further enhance agent's reasoning capacity, and enable it to periodically learn from previous failed experience.

\subsubsection{Reasoning Format Cold Start with SFT}
We first introduce our data synthesis method and then describe the fine-tuning objective.

\textbf{Cold-Start Data Synthesis and Filtering.}
We distill multi‑turn trajectories from DeepSeek‑R1 under the sandbox, yielding complete traces $y=\{\tau_{1},a_{1},o_{1},\ldots,\tau_{t},a_{t}\}$ that interleave thoughts, tool calls, tool responses, and final answer. 
Thoughts $\tau_i$ are wrapped by \thinkstart{}\ldots\thinkend{}, actions $a_i$ are either function calls enclosed by \tollcallstart{}\ldots\tollcallend{} or the final itinerary answer $a_t$ enclosed by \answerstart{}\ldots\answerend{}, and observations $o_i$ are tool responses enclosed by \tollresponsestart{}\ldots\tollresponseend{}.
Then, we utilize the constructed reward modeling system to filter incorrect trajectory, and finally we apply strict format checks to retain only sequences correctly segmented by the special tags.

\textbf{Training Objective.}
We train the TP agent based on the above verified and filtered dataset $\mathcal{D}$. 
The instruction input concatenates a system prompt $T$ with the user query $q$, and the output is the verified trajectory $y$:
\begin{equation}
\label{fine-tune}
\begin{aligned}
\mathcal{L} = \mathbb{E}_{(q, y) \sim \mathcal{D}}\left[\log \mathcal{P}_{\pi_{\theta}}(y \mid q)\right],
\end{aligned}
\end{equation}
where $\mathcal{L}$ represents the loss function and $\pi_{\theta}$ is the agent policy.
In practical training process, the tokens corresponding to the agent’s environmental observations $o_{i}$ are masked out from the loss calculation \cite{jin2025search}.
The system prompt could be found in Appendix \ref{appendix_prompt}.


\subsubsection{Reinforcement Learning with Experience Replay}
After cold-start, we derive a two-phase process that first saves verified unsuccessful trajectories as a query buffer and then replays them in subsequent training steps \cite{zhang2025rlep}.

\textbf{Rollout and Replay Strategy.}
Following the sampling procedure in Group Relative Policy Optimization (GRPO) \cite{shao2024deepseekmath}, we sample a group of trajectories $\left \{y_1, y_2, .., y_n\right \}$ for each query $q$.
If none of the trajectories in the group yields a verified correct answer, we store the query in an experience buffer $B$ for later replay.
The motivation is that, after subsequent RL training steps, the improved policy may generalize to handle previous failed hard sample \cite{xie2025logic}.

\textbf{Policy Optimization.}
Set RL training dataset as $D$, experience buffer as $B$, which is replayed after fixed training step $\gamma $. 
We formulate the optimization goal as follows:
\begin{equation}
\begin{aligned}
& \max_{\pi_{\theta}}\mathbb{E}_{\substack{q \sim \left\{D, B\right\},\left\{y_{i}\right\}_{i=1}^{n} \sim \pi(y \mid q)}} \Bigg[
\frac{1}{n} \sum_{i=1}^{n}\Big(\min \Big(\frac{\pi_{\theta}\left(y_{i} \mid q\right)}{\pi_{ref}\left(y_{i} \mid q\right)} A_{i}, \\
& \quad \operatorname{clip}\Big(\frac{\pi_{\theta}\left(y_{i} \mid q\right)}{\pi_{ref}\left(y_{i} \mid q\right)}, 1-\varepsilon, 1+\varepsilon\Big) A_{i}\Big)\Big) -\beta \mathbb{D}_{\mathrm{KL}}\left[\pi_{\theta} \| \pi_{\mathrm{ref}}\right]\Bigg]
\end{aligned}
\end{equation}
where $\varepsilon$, $\beta$ are hyperparameters, $n$ is the rollout size, $\mathbb{D}_{\mathrm{KL}}$ denotes the KL-divergence, and $A_{i}=r_i-avg(r)/std(r)$ represents the advantage, which is computed based on the group rewards $r=\left \{r_1, r_2, .., r_n\right \}$.
In this work, we propose to filter out samples when the standard deviation of group rewards satisfies $std(r)\le \eta $, where $\eta$ is set to 0.1. 
This strategy aims to exclude samples that are either too simple or too hard, where the agent receives similar rewards even under large-size rollouts, thereby encouraging more effective exploration of the current policy.
In addition, we utilize loss masking operation for tool responses tokens (wrapped with \tollresponsestart{} and \tollresponseend{}) to ensure policy gradient is computed only over agent-generated tokens.

\section{System Deployment}
\begin{figure}
    \centering
    \includegraphics[width=1\linewidth]{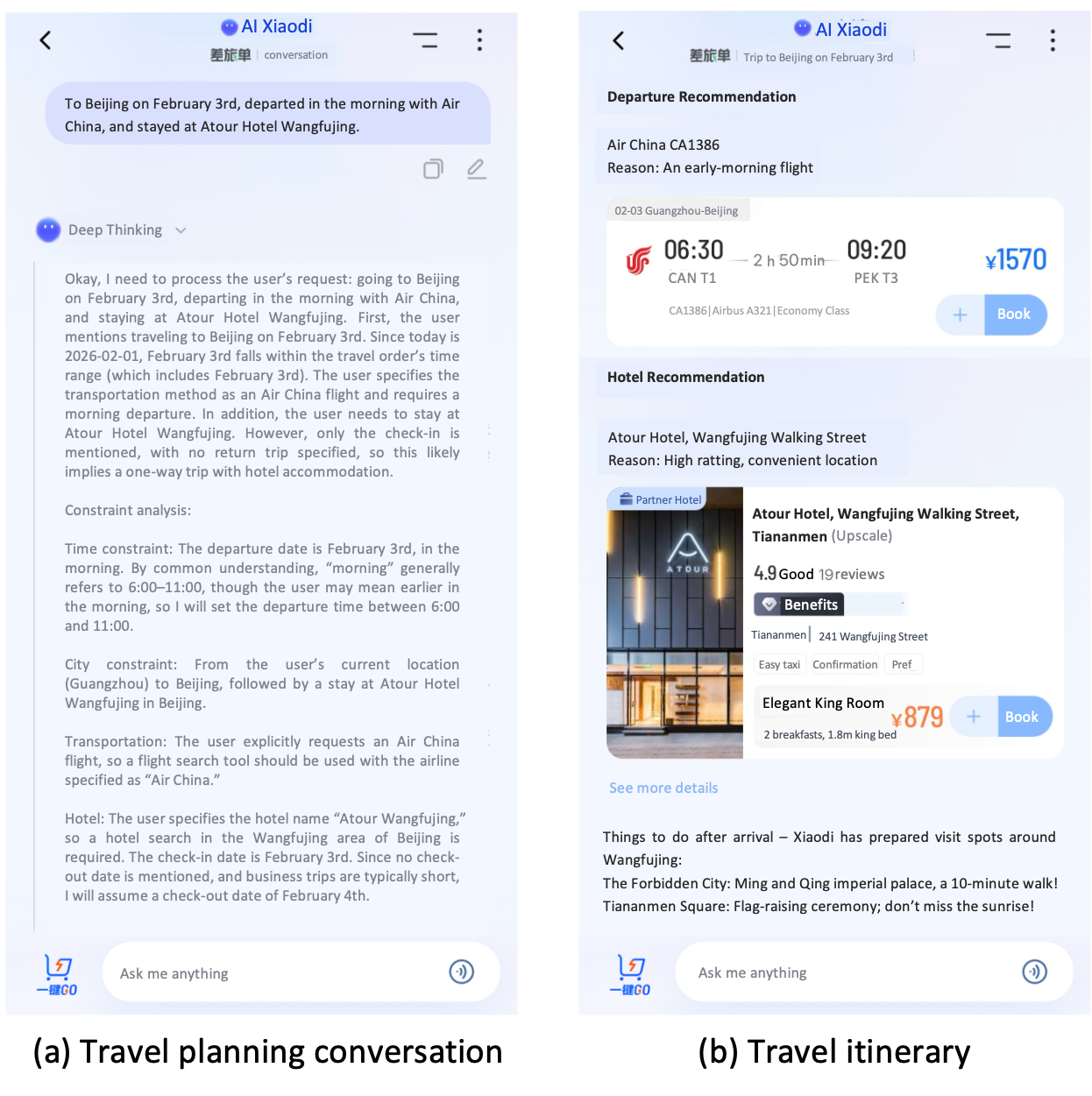}
    \caption{The conversational interface of DeepTravel in DiDi Enterprise Solution App.}
    \label{fig_demo}
    \vspace{-10pt}
\end{figure}

Since June 2025, DeepTravel has been fully deployed into the DiDi Enterprise Solution (ES) application, providing an intelligent, agent-based travel planning service shown in Figure \ref{fig_demo}. 
In this section, we detail the deployment strategy, focusing on model adaptation from large-size reasoning LLMs, the asynchronous update loop, and the online serving infrastructure.

\textbf{Backbone selection and adaptation.}
In the preliminary phase, we utilized DeepSeek-R1 (671B) as the agent backbone to prototype the travel planning service on the DiDi ES App. However, we observed that DeepSeek-R1 suffered from suboptimal performance due to a lack of domain-specific training, alongside prohibitive inference latency caused by its 671B parameter size. 
Consequently, we transitioned to Qwen3-32B as the backbone and applied the proposed DeepTravel framework to conduct SFT and large-scale reinforcement learning. 
For the SFT cold-start, we use about 1K training samples, with a learning rate of 5e-6, over 2 epochs.
For RL training, we select roughly 500 high-quality samples and set the rollout size to 8 and use a learning rate of 5e-7.
Additionally, the maximum sequence length of agent is set as 32K tokens, and the maximum interaction turns is limited to 8.
The training of Qwen3-32B requires 32 H800 NVIDIA GPUs, and the training process takes approximately 30 hours per hundreds steps.
This process distilled the reasoning capabilities into a more compact architecture. 
The resulting DeepTravel-32B was deployed to our production environment, achieving superior service quality with significantly reduced token cost compared to the initial prototype.

\textbf{Asynchronous agent update.}
To address emerging corner cases and evolving user needs, we established a asynchronous agent update mechanism. 
A dedicated professional annotation team, in collaboration with product managers, monitors the online environment to identify and label service failures (bad cases).
Based on these identified issues, we employ a dual-strategy for refinement: (1) optimizing system prompts (e.g., add stricter rubrics), or (2) integrating the failure cases into the SFT and RL datasets for model retraining. 
Updated models undergo rigorous validation by the product team before being pushed to the online environment, ensuring stability and continuous improvement.

\textbf{Online serving strategy.}
We deploy DeepTravel-32B on a cluster consisting of two nodes, each equipped with 4 NVIDIA H800 GPUs. 
In the production environment, DeepTravel-32B generates a complete travel itinerary with an average latency of approximately 25 seconds. 
Decomposing this latency shows that the model spends approximately 19 seconds on the “deep thinking” process (Figure~\ref{fig_demo}(a)) and 6 seconds on answer summarization and frontend order-page rendering (Figure~\ref{fig_demo}(b)). 
This optimized deployment strategy represents a substantial system efficiency gain, reducing the average inference latency by approximately 90 seconds compared to the direct utilization of DeepSeek-R1, thereby ensuring a responsive user experience.


\begin{table*}
\centering
\caption{Datasets statistics. We categorize the data into the Training Set (including SFT and RL stages) and Testing Set (Online production data and synthesized Offline data).}
\label{dataset}
\begin{tabular}{l|c c|c|c c c|c c c}
\toprule
\multirow{4}{*}{\textbf{Statistics}} 
& \multicolumn{2}{c|}{\textbf{Training Set}} 
& \multicolumn{7}{c}{\textbf{Testing Set}} \\
& \multirow{3}{*}{\textbf{SFT}} 
& \multirow{3}{*}{\textbf{RL}} 
& \multirow{3}{*}{\textbf{Online}} 
& \multicolumn{6}{c}{\textbf{Offline}} \\
& 
& 
& 
& \multicolumn{3}{c|}{\textbf{with constraint}} 
& \multicolumn{3}{c}{\textbf{without constraint}} \\
& 
& 
& 
& Easy & Medium & Hard 
& Easy & Medium & Hard \\
\midrule
Time span 
& - 
& - 
& 2025/06/01 -- 2025/08/31 
& \multicolumn{3}{c}{-} & \multicolumn{3}{c}{-}\\
\# of queries 
& 1,086 
& 500 
& 6,224 
& 222 & 78 & 200 
& 156 & 45 & 299 \\
\bottomrule
\end{tabular}
\end{table*}

\section{Experiments}
We conduct extensive experiments to evaluate the proposed method and aim to answer the following research questions: 
\begin{itemize}[leftmargin=*, nosep]
\item \textbf{RQ1}: How does DeepTravel perform in the online environment? 
\item \textbf{RQ2}: How does DeepTravel compare with existing state-of-the-art approaches (e.g., frontier LLM agents, prior travel planning methods, and RL algorithms)?
\item \textbf{RQ3}: How do different components (e.g., cold-start, hierarchical reward, experience replay) affect the DeepTravel’s performance? 
\item \textbf{RQ4}: How does DeepTravel behave (e.g., model entropy and sandbox stability) during agentic RL training? 
\item \textbf{RQ5}: How about DeepTravel's real-world user experience?
\end{itemize}

\subsection{Experimental Setup}

\subsection{Data Curation}

\textbf{Data Synthesis and Training Sets.} 
To construct user queries, we systematically enumerate combinations of atomic intents (e.g., origin, destination, departure time, and budget) and inversely generate natural language user queries. We then perform repeated sampling using large language models (e.g., Qwen-3-32B and DeepSeek-R1). Each candidate query is evaluated by our reward modeling system, which assigns a difficulty score. For \textbf{SFT Cold-Start}, we prioritize relatively simple queries while retaining a small proportion of difficult ones to improve model robustness, resulting in approximately 1K filtered trajectories. For \textbf{RL Training}, we conversely emphasize complex queries but include a small fraction of simple cases to ensure coverage. To ensure stable optimization, we incorporate a strict manual inspection stage to remove semantically invalid or ill-posed queries. This yields 500 high-quality, human-annotated samples for RL (split into 450 for training and 50 for validation).

\textbf{Online and Offline Evaluation.} 
Our framework is evaluated in both online and offline environments. For \textbf{offline testing}, we synthesize two balanced categories based on explicit user personalization constraints (e.g., budget restrictions or individual travel preferences): 500 queries \textit{with constraints} and 500 queries \textit{without constraints}. Product experts annotate the complexity of each query based on human preference, categorizing them into easy, medium, and hard levels. For \textbf{online testing}, we collected 6,224 real-world queries from the online production environment of the DiDi ES App over a three-month period (June 1st to August 31st). Table \ref{dataset} summarizes the dataset statistics.

\begin{table}
    \centering
    \caption{Alignment analysis between Human Evaluation and LLM-as-a-Judge. The results are based on 200 samples.}
    \label{tab:human_llm_alignment}
    \scalebox{0.9}{
    \begin{tabular}{l c c}
    \toprule
    \textbf{Evaluation protocol} & \makecell{\textbf{Raw Accuracy} $\to$ \\ \textbf{Double-check Accuracy}} & \makecell{\textbf{Agreement}\\ \textbf{Error}} \\
    \midrule
    Human evaluation & 100\% $\to$ 93\%  & \multirow{2}{*}{4\%} \\
    LLM-as-a-Judge & 82\% $\to$ 89\% & \\
    \bottomrule
    \end{tabular}}
    \vspace{-5pt}
\end{table}

\begin{table}

    \centering
    \caption{Online performance of DeepTravel-32B across seven human annotation dimensions.}
    \label{tab:deeptravel_performance}
    \begin{tabular}{lc}
        \toprule
        \textbf{Evaluation dimension (\%)} & \textbf{DeepTravel} \\
        \midrule
        User intent understanding & 92.85 \\
        Itinerary completeness & 98.04 \\
        Itinerary feasibility & 87.63 \\
        Itinerary affordability & 84.42 \\
        Itinerary clarity & 92.16 \\
        Personalized requirement & 76.62 \\
        Without factual hallucination & 84.42 \\
        \midrule
        \textbf{Final Pass Rate} & \textbf{82.58} \\
        \bottomrule
    \end{tabular}
        \vspace{-5pt}
\end{table}

\subsubsection{Metric and Evaluation Protocol.}
We adopt the Final Pass Rate from TravelPlanner \cite{xie2024travelplanner} as our primary evaluation metric. 
This metric directly assesses the validity of a generated travel itinerary. 
Regarding the evaluation protocol, we employ both of LLM-as-a-Judge and human evaluation.

\textbf{For LLM-as-judge}, We apply our custom reward modeling system to determine whether a generated itinerary is correct.
\textbf{For Human Evaluation}, we decompose the assessment into seven distinct dimensions: intent understanding, itinerary completeness, feasibility, affordability, clarity, satisfaction of personalized requirements, and factual hallucination. 
An itinerary is correct only if it satisfies the criteria across all seven rubrics.
In our experiments, we utilize LLM-as-a-judge for full-sample evaluation. 
Additionally, for each evaluation set, we randomly sample 200 instances for human verification to assess the reliability of our automated evaluator. 

\textbf{Evaluation Quality Analysis.}
Overall, the raw agreement between LLM-as-a-Judge and human evaluation is 82\%.
However, Table \ref{tab:human_llm_alignment} highlights that human annotation carries a 7\% error rate. 
By accounting for this inherent label noise, the LLM’s actual evaluation accuracy improves to approximately 89\%. 
This minimal marginal gap ($\sim$4\%) demonstrates that the LLM is well-aligned with human's "gold" standard, justifying its use as a reliable and scalable proxy for large-scale agentic RL training and evaluation.
We provide more detailed evaluation rubrics description in Appendix \ref{appendix_annotation}.

\begin{table*}
\caption{Overall Final Pass Rate (\%) results on synthesized offline travel planning benchmarks. The best results are \textbf{bolded}, and the best baseline results in each setting are \underline{underlined}.}
\label{main_results_merged}
\centering
\begin{tabular}{l|c|ccc|ccc|c}
\midrule
\multirow{3}{*}{\textbf{Category}} & \multirow{3}{*}{\textbf{Model}} & \multicolumn{6}{c|}{\textbf{Offline}} & \multirow{3}{*}{\begin{tabular}[c]{@{}c@{}}\textbf{Human}\\ \textbf{Evaluation}\end{tabular}} \\
& & \multicolumn{3}{c|}{\textbf{Without constraint}} & \multicolumn{3}{c|}{\textbf{With constraint}} & \\
& & \textbf{Easy} & \textbf{Medium} & \textbf{Hard} & \textbf{Easy} & \textbf{Medium} & \textbf{Hard} & \\
\midrule
\multirow{9}{*}{\begin{tabular}[l]{@{}l@{}}Frontier \\ Reasoning \\ LLMs\end{tabular}} 
& DeepSeek-R1 & 45.55 & 34.74 & \underline{26.00} & \underline{65.36} & 43.33 & \underline{27.09} & \underline{72.00} \\
& OpenAI-o1 & 36.57 & 33.16 & 20.60 & 30.36 & 24.44 & 17.69 & 54.00 \\
& OpenAI-o3 & 37.30 & 20.11 & 21.19 & 37.50 & 26.67 & 15.69 & 52.00 \\
& K2 & \underline{54.01} & \underline{48.42} & 25.52 & 57.14 & \underline{53.33} & 21.40 & 64.00 \\
& Qwen3-235B & 38.69 & 36.84 & 20.24 & 44.64 & 26.67 & 10.37 & 52.00 \\
& gpt-oss-120B & 40.15 & 27.37 & 20.83 & 64.29 & 42.22 & 16.39 & 48.00 \\
& Seed-OSS-36B & 23.65 & 13.16 & 11.19 & 25.00 & 13.33 & 12.34 & 20.00 \\
& Qwen3-32B & 29.85 & 27.89 & 23.21 & 53.57 & 25.00 & 9.03 & 38.00 \\
& Qwen3-8B & 10.95 & 9.47 & 4.76 & 28.57 & 26.67 & 5.35 & 26.00 \\ 
\midrule
\multirow{2}{*}{\begin{tabular}[l]{@{}l@{}}Existing TP \\ Frameworks\end{tabular}} 
& DeepSeek-R1 + TripTailor & 43.37 & 38.74 & 18.53 & 63.32 & 40.56 & 24.47 & 66.00 \\
& DeepSeek-R1 + TravelAgent & 39.95 & 42.28 & 23.34 & 63.36 & 40.14 & 22.26 & 68.00 \\
\midrule
\multirow{3}{*}{\begin{tabular}[l]{@{}l@{}}Frontier RL \\ Frameworks\end{tabular}} 
& Qwen3-8B + PPO & 48.26 & 33.25 & 14.62 & 60.05 & 34.86 & 15.04 & 62.00 \\
& Qwen3-8B + GRPO & 52.36 & 34.06 & 13.52 & 61.78 & 36.65 & 15.82 & 64.00 \\
& Qwen3-8B + DAPO & 52.06 & 35.52 & 15.04 & 62.24 & 40.02 & 16.54 & 64.00 \\ 
\midrule
\multirow{4}{*}{Ours} 
& DeepTravel-8B-Cold-Start & 41.09 & 31.58 & 12.64 & 56.07 & 28.89 & 12.37 & 58.00 \\
& DeepTravel-8B-RL & 54.25 & 36.84 & 20.24 & 64.86 & 41.89 & 21.40 & 70.00 \\
& DeepTravel-32B-Cold-Start & 56.42 & 32.95 & 25.60 & 61.07 & 40.44 & 17.52 & 66.00 \\ 
& DeepTravel-32B-RL & \textbf{69.34} & \textbf{54.74} & \textbf{29.17} & \textbf{73.21} & \textbf{62.22} & \textbf{35.75} & \textbf{82.00} \\
\midrule
\end{tabular}
        \vspace{-5pt}
\end{table*}
\subsubsection{Baselines.}
We first compare our method to nine reasoning LLMs, each is derived as a TP agent under the same framework.
These baselines include DeepSeek-R1 \cite{guo2025deepseek}, OpenAI-o1 \cite{jaech2024openai}, OpenAI-o3, K2 \cite{team2025kimi}, Qwen3-235B, gpt-oss-120B \cite{openai2025gptoss120bgptoss20bmodel}, Seed-OSS-36B \cite{seed2025seedoss}, Qwen3-32B, and Qwen3-8B \cite{yang2025qwen3}.
In addition, we also compare DeepTravel with two existing travel planning framework, e.g., TripTailor \cite{wang2025triptailor} which employs a prompting-based ReAct framework and TravelAgent \cite{chen2024travelagent}, which utilizes a multi-agent workflow for itinerary generation.
Finally, we compare our proposed DeepTravel framework with three representative on-policy RL training methods, i.e., PPO \cite{schulman2017proximal}, GRPO \cite{shao2024deepseekmath} and DAPO \cite{yu2025dapo}.

\subsection{Online Test (RQ1)}
Table \ref{tab:deeptravel_performance} summarizes the human annotation results of DeepTravel in the online production environment. 
Overall, the model achieves a Final Pass Rate of 82.58\%. 
It excels in understanding user intent, itinerary completeness, and clarity, with all three dimensions scoring above 92\%. 
However, performance drops in personalized requirement (76.62\%) and factual hallucination issue (84.42\%). 
These results indicate that DeepTravel-32B is highly effective at structuring coherent travel itinerary.
However, DeepTravel still faces significant challenges in satisfying complex personal preferences and avoiding factual hallucinations.

\begin{table*}[]
\caption{Ablation study of cold-start and RL on Qwen3-8B.
}
\label{ablation_study_results}
\centering
\begin{tabular}{c|ccc|ccc|c}
\midrule
\multirow{3}{*}{\textbf{Model variants}} & \multicolumn{6}{c|}{\textbf{Offline}} & \multirow{3}{*}{\begin{tabular}[c]{@{}c@{}}\textbf{Human}\\ \textbf{Evaluation}\end{tabular}}\\
& \multicolumn{3}{c|}{\textbf{Without constraint}} & \multicolumn{3}{c|}{\textbf{With constraint}} & \\
 & \textbf{Easy} & \textbf{Medium} & \textbf{Hard} & \textbf{Easy} & \textbf{Medium} & \textbf{Hard} & \\
\midrule
DeepTravel-8B w/o ER & 51.01 & 32.21 & 8.81 & 60.86 & 35.00 & 8.75 &  66.00 \\
DeepTravel-8B w/o CS & 45.99 & 25.26 & 16.79 & 53.57 &35.56 & 22.18 &  48.00\\ 
DeepTravel-8B w/o Traj & 50.26 & 35.47 & 18.24 & 61.06 & 33.25 & 20.75 & 66.00\\
DeepTravel-8B w/o Turn & 52.05 & 28.04 & 5.25 & 59.04 &14.24 & 10.76 &  58.00\\ 

\midrule
DeepTravel-8B-RL & \textbf{54.25} & \textbf{36.84} & \textbf{20.24} & \textbf{64.86} & \textbf{41.89} & \textbf{21.40} & \textbf{70.00}\\
\midrule
\end{tabular}
        \vspace{-5pt}
\end{table*}

\begin{figure*}
    \centering
    \includegraphics[width=1\linewidth]{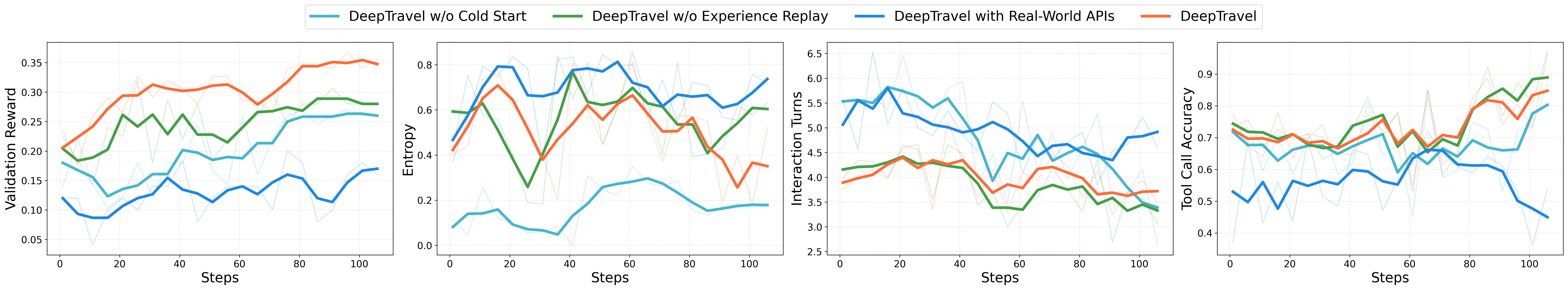}
    \caption{Validation reward (final pass rate), model entropy, average interaction turn and tool call accuracy (success rate) throughout agentic RL training process.}
    \label{training_statistics}
        \vspace{-5pt}
\end{figure*}
\subsection{Offline Test (RQ2)}
As reported in Table \ref{main_results_merged}, we compare the final pass rate of DeepTravel with existing state-of-the-art reasoning LLMs and RL frameworks. 

\begin{table*}[]
\centering
    \caption{Statistics of the constructed travel sandbox and agent tool call count during one epoch of agentic RL training.}
    \label{tab:sandbox_statistics}
    \scalebox{0.9}{
\begin{tabular}{ccccccccc}
\midrule
    \textbf{Type}   & \textbf{Total Mock data} & \textbf{Flight search} & \textbf{Train search} & \textbf{Route planning} & \textbf{Hotel search} & \textbf{POI search} & \textbf{Web search} & \textbf{Total tool call} \\ \midrule
\# of num & 117,844         & 899           & 1,835        & 3,267          & 1,211        & 2,657      & 3,014 & 12,883      \\ \midrule
\end{tabular}
}
        \vspace{-5pt}
\end{table*}

\textbf{Comparison with Existing Reasoning LLMs.}
We compare DeepTravel across the SFT cold-start stage and RL training process.
Overall, DeepTravel achieves significant improvement compared with the the state-of-the-art reasoning agents using both of online and offline evaluation setting.
In addition, We highlight two key observations:
\textbf{(i) DeepTravel substantially boosts small-size LLMs}. 
For instance, Qwen3-8/32B is improved to the state-of-the-art levels, matching and even surpassing more heavily and much larger frontier LLMs.
On offline without constraint setting, DeepTravel-8B and DeepTravel-32B achieves an final pass rate of 54.25\% and 69.34\%, outperforming K2 by 0.1\% and 28.9\%, respectively.
For other setting, DeepTravel-8B achieves comparable performance and DeepTravel-32B consistently outperforms frontier reasoning LLMs, such as DeepSeek-R1, OpenAI-o1 and OpenAI-o3.
\textbf{(ii) The agentic RL training continually improves domain-specific reasoning capacity}.
As reported, while cold-start stage could establish a strong initial policy compared to base model, the following agentic RL  yields surprisingly performance improvement.
Specifically, agentic RL further boost initial cold-start policy of DeepTravel-8B and DeepTravel-32B by 24\% (40.00 to 49.75) and 25.5\% (50.03 to 62.77) on online experimental setting, respectively.
The improvement in the offline setting is also significant.

\textbf{Comparison with Existing TP framework and RL Algorithms.}
As reported in Table \ref{main_results_merged}, existing powerful travel planning frameworks (e.g., TripTailor and TravelAgent), even when using DeepSeek-R1 as the backbone, achieve significantly lower performance than DeepTravel-32B-RL.
This demonstrate the advantage of DeepTravel framework.
In addition, we also compare DeepTrave with three recent popular reinforcement learning methods on DeepTravel-8B-Cold-Start backbone.
For each RL methods, we run 100 training steps with the same training sample.
As can be seen, While all online policy learning methods show improvement over the base model, our replay-augmented RL approach demonstrates superior performance, especially when tackling hard queries.

\subsection{Ablation Study (RQ3)}
To validate the effectiveness of each module in DeepTravel, we conduct an ablation study on the Qwen3-8B dataset.
Specifically, we compare the following variants.
(1) DeepTravel-8B w/o ER removes the \textbf{E}xperience \textbf{R}eplay module in RL training process.
(2) DeepTravel-8B w/o CS removes the SFT-based \textbf{C}old \textbf{S}tart stage before conduct reinforcement learning.
(3) DeepTravel-8B w/o Traj removes the \textbf{Traj}ectory-Level verifier in reinforcement learning training process.
(4) DeepTravel-8B w/o Turn removes the \textbf{Turn}-Level verifier in reinforcement learning training process.

As shown in Table \ref{ablation_study_results}, we obtain the following observations.
First, the experience replay strategy is important for the training.
Removing it will decrease model performance.
Second, the cold-start stage seems to be critical for RL training as we obtain significant performance decrease after removing it.
The potential reason lies on that the cold-start will help LLMs learn basic tool usage, instruction following capacity.
Finally, we observe that the turn-level verifier contributes more to the model performance.
When removing it, agent's performance decreases and it performs poorly on hard problem.
The reason may lie that more complex problem requires verification turn-by-turn.
However, the trajectory-level verifier also proves important, as its removal causes a performance decline as well. 
In addition to its contribution to accuracy, the trajectory-level verifier enhances training efficiency by removing the need of fine-grained turn-level verification.

\begin{figure*}
    \centering
    \includegraphics[width=1\linewidth]{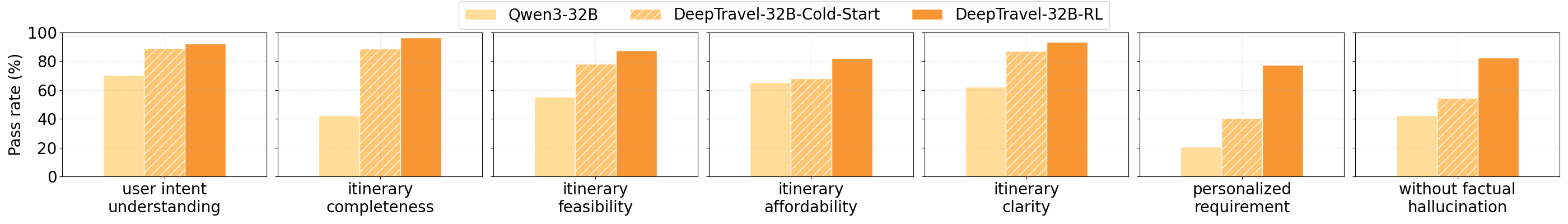}
    \caption{Capacity comparison of the autonomous TP agent across 7 human-annotated dimensions, evaluated on 50 randomly sampled real-world online user case.}
    \label{annotation_dimension}
    \vspace{-10pt}
\end{figure*}

\subsection{In-Depth Analysis (RQ4)}
This section presents an in-depth analysis on cold-start, experience replay strategy and sandbox construction.
Specifically, we present the validation reward, model entropy, average interaction turns, and tool-call accuracy throughout the agentic RL training process in Figure~\ref{training_statistics}. 
We highlight the following findings:

\textbf{Impact of Cold-Start and Experience Replay Strategies.}
The cold-start strategy helps initialize a reasonably effective policy, particularly in terms of producing a more appropriate number of tool-interaction turns (whereas the base model tends to overuse tool calls). The experience replay strategy contributes little in the very early training stages. 
However, by progressively replaying previously failed samples, it steadily enhances the model’s capacity and eventually leads to substantial improvements over the no-replay baseline in later training steps.

\textbf{Impact of Travel Sandbox.}
We compare agentic RL training with real-world APIs and with the constructed sandbox. 
As shown in Figure \ref{training_statistics}, tool-call accuracy with real-world APIs is unstable and consistently lower than that achieved in the sandbox. 
Under these circumstances, the TP agent shows no clear reward improvement, highlighting the importance of a stable sandbox environment for agentic Rl training.
In addition, we also provide detailed tool call statistics of the constructed sandbox during a single RL training epoch using 500 queries.
As presented in Table \ref{tab:sandbox_statistics}, the sandbox caches about one million mock data, ensuring stable agent RL training.
Supported by this stable sandbox environment, the agent successfully executed totally 12,883 tool calls within a single training run, demonstrating the robustness of the sandbox.

\textbf{Non-decreasing Entropy in Agentic RL.}
We observe a non-decreasing entropy phenomenon during agentic RL training, which is also posed by several very recent studies \cite{dong2025agentic}.
We think the behind reason lies on that the TP agent need to continually adapt its policy to the dynamically changing responses of external tools.

\subsection{Real-World User Study (RQ5)}
We conducted a real-world user study based on evaluation dimensions defined by DiDi’s ES product manager and annotation team, with the results summarized in Figure \ref{annotation_dimension}. 
Overall, both the cold-start and RL approaches improved user satisfaction across the seven evaluation dimensions. 
In particular, the cold-start method substantially enhanced the model’s fundamental capabilities, especially in understanding user intentions and in improving the completeness, feasibility, and clarity of the generated travel itineraries. 
However, for more advanced capabilities—such as capturing and satisfying personalized preferences—the cold-start approach alone proved insufficient, suggesting that these aspects may require large-scale exploration during the RL stage. 
Finally, we observed that both the base model and the cold-start model suffered from severe hallucination issues, with factual error rates reaching up to 50\%. 
RL training is able to effectively address this problem, reducing hallucinations to below 20\%.
More annotation insights is in Appendix \ref{appendix_annotation}

\section{Related Work}
\subsection{LLMs as Travel Planning Agent.}
Large language models (LLMs) have reshaped travel planning (TP) by enabling LLM-powered TP agent to interact with external tools for itinerary generation that aligns with user preferences.
In the literature, two major paradigms have emerged to construct TP agent: (i) hand-craft prompt tuning, and (ii) fixed agent workflow design.
Hand-craft prompt tuning approaches \cite{shao2025personal}—including TravelPlanner \cite{xie2024travelplanner}, TripTailor \cite{wang2025triptailor}, and ChinaTravel \cite{shao2024chinatravel}—decompose the end-to-end task into multiple sub-steps \cite{ni2025tp} and introduce tailored evaluation metrics for each stage. 
While effective, their practical value is limited by weak grounding to dynamic real-world environment (e.g., tool availability).
To this end, recent work integrates prompt strategies into well-structured agent pipelines. 
Representative studies include TravelAgent \cite{chen2024travelagent}, PTS \cite{shao2025personal} and RETAIL \cite{deng2025retail} design fixed workflows to enhance intention understanding, orchestrate external tools, and ensure end-to-end itinerary generation.
However, they are still labor‑intensive to build and maintain, and they generalize poorly to new user queries or changing tools and resources, limiting the flexibility and autonomy of TP agents.

\subsection{Agentic Reinforcement Learning for LLMs.}
Agentic reinforcement learning (RL) has recently been widely applied across domains to build autonomous AI agents \cite{zhang2025landscape}, wherein the agent interacts with tools in a dedicated sandbox environment and iteratively improves its policy based on received reward feedback \cite{shang2025rstar2}.
For instance, ReTool \cite{feng2025retool}, Kimi-Researcher \cite{kimiResearcher} and WebSailor \cite{li2025websailor} are constructed reasoning agent in math, deep research and web domain.
In addition, many recent work like rStar2-Agent \cite{shang2025rstar2} and AgentGym-RL \cite{xi2025agentgym} make attempts to propose a unified agentic RL training framework across diverse domains, facilitating the construction of foundation agent.
Nevertheless, the application of agentic RL in travel planning domain remains unexplored.

\section{Conclusion and Future Work}
In this work, we propose DeepTravel, the first end-to-end agentic RL training framework to build autonomous travel planning (TP) agent, offering new paradigm for current TP studies.
We first construct a robust sandbox, where the TP agent could be trained without limitation of real-world APIs issues (e.g., QPS limits and inconsistency outputs).
Then, we propose a hierarchical reward modeling system, which first devise a coarse-grained trajectory verifier for high-level spatiotemporal requirement verification, and then use a turn-level verifier to verify agent's answer step-by-step.
Finally, we propose a replay-augmented reinforcement learning alogrithm, which allow travel agent to periodically replay previous failed case, improve its out-of-domain generalization capacity.
We deploy the resulted autonomous TP agent in DiDi ES App, demonstrating the deployment value of DeepTravel. 
Extensive experiments on online production environment and offline synthetic data show that DeepTravel enable small-size LLMs to significantly outperform frontier reasoning LLMs and TP agent framework.
In the future, we aim to develop a more flexible and automatic reward model and extend this framework to other domains.

\section*{Acknowledgments}
This work was supported by the National Natural Science Foundation of China (Grant No. 62572417, No.92370204), National Key R\&D Program of China (Grant No.2023YFF0725004), and CCF-DiDi GAIA Collaborative Research Funds.
In the deployment of the DeepTravel, we would like to thank Jia Chen, Xuzhong Huang, Qicheng Hu, Han Li, Pengcheng Li, Lipeng Liang, Qingyi Liang, Yang Liu, Jiyuan Ren, Shengqing Zhai, and Yubo Zhou from Didichuxing Co. Ltd for their participation in agent RL post-training.
The names are listed in alphabetical order by last name.

\bibliographystyle{ACM-Reference-Format}
\bibliography{sample-base}

\appendix

\section{Appendix}

\subsection{Training Details}
\label{appendix_training}
For SFT-based cold-start, we use Megatron-LM for fine-tuning, and build our RL training framework on top of verl.
In the RL training process, we select DeepSeek-R1 as the backbone of our verifier.
For an 8B-parameter LLM, all training can be completed on a single node with 8 H800 GPUs.
For a 32B-parameter LLM, training requires 32 H800 GPUs across 4 nodes.
SFT takes about 2–3 hours for 1,000 training samples. 
RL takes about 30 hours per 100 steps for 500 samples.

\begin{table*}
\centering
\caption{Evaluation dimensions for TP agent on human annotation.}
\label{tab:evaluation_dimensions}
\small
\setlength{\tabcolsep}{4pt}
\begin{tabular}{
>{\raggedright\arraybackslash}p{3.1cm}
>{\raggedright\arraybackslash}p{4.8cm}
c
>{\raggedright\arraybackslash}p{5.6cm}
}
\toprule
\textbf{Dimension} & \textbf{Explanation} & \textbf{Score} & \textbf{Criteria} \\
\midrule

User Intention Understanding
& Correctly identifies user requirements such as departure/destination, time, budget, and travel scenario
& T/F
& \textbf{T}: All key elements correctly identified; 
\textbf{F}: Missing or incorrect elements. \\

Itinerary Completeness
& Covers key components including flights, hotels, and transportation with coherent timing
& T/F
& \textbf{T}: Complete and coherent itinerary; 
\textbf{F}: Missing components or inconsistent timing/location. \\

Itinerary Feasibility
& Whether the itinerary is practically executable
& T/F
& \textbf{T}: Reasonable and executable arrangement; 
\textbf{F}: Tight schedules or unreasonable combinations. \\

Itinerary Affordability
& Whether resources are bookable, accurately priced, policy-compliant, and cost-effective
& T/F
& \textbf{T}: Real and policy-compliant options; 
\textbf{F}: Unbookable or overpriced options. \\

Itinerary Clarity
& Whether recommendations are generated efficiently with clear justification
& T/F
& \textbf{T}: Fast response with understandable reasoning; 
\textbf{F}: Slow response or vague explanation. \\

Personalized Requirement
& Whether recommendations consider user history and preferences
& T/F
& \textbf{T}: Matches user preferences; 
\textbf{F}: Ignores personalized requirements. \\

Without Factual Hallucination
& Whether the recommendation contains fabricated information
& T/F
& \textbf{T}: Factually correct recommendations; 
\textbf{F}: Hallucinated or logically inconsistent content. \\

\bottomrule
\end{tabular}
\end{table*}
Below we detail the supervision signals and diagnostics we track to ensure stable and effective training of DeepTravel. 
For the TP agent, we continuously monitor entropy, gradient norm, average response length, reward, and average turn count. 
We also track metrics tied to the broader RL loop—tool-call accuracy and external verifier success rate—to capture the influence of the sandbox environment, the reward modeling system, and the RL algorithm itself. 
For the RL method, we additionally log the sample keep rate and the loss-mask ratio throughout training. 
These metrics jointly inform training stability and failure modes: low entropy suggests poor exploration, while excessively high gradient norms indicate instability; unusually short responses and few turns often signal reward hacking; low tool-call accuracy and verifier success point to systematic execution or evaluation errors; and a very low sample keep rate typically means the data regime is misaligned (too easy or too hard), reducing the need—or opportunity—for exploration.


\subsection{Human Annotation}
\label{appendix_annotation}
In this section, we provide detailed description of how the human annotation pipeline works.
As shown in Table \ref{tab:evaluation_dimensions}, the human annotation process consists of seven dimensions designed to comprehensively assess the quality of AI-powered travel itinerary:
\begin{itemize}
\item \textbf{User Intention Understanding} evaluates the system's ability to correctly parse and interpret user inputs, ensuring all critical travel parameters are accurately captured. This dimension is fundamental as misunderstanding user requirements leads to irrelevant recommendations.

\item \textbf{Itinerary Completeness} assesses whether the recommendation covers all essential travel components (flights, accommodation, local transportation) and maintains temporal coherence. A complete itinerary should provide seamless transitions between different travel segments.

\item \textbf{Itinerary Feasibility} examines the practical executability of the proposed itinerary. This includes verifying that the schedule is not overly ambitious, transportation connections are realistic, and the overall route forms a logical closed loop.

\item \textbf{Itinerary Affordability} focuses on the economic and practical aspects of the recommendation, ensuring resources are actually bookable, prices are accurate, and recommendations comply with organizational travel policies while maintaining cost-effectiveness.

\item \textbf{Itinerary Clarity} measures both the efficiency of the reasoning process and the transparency of the reasoning provided. 
Quick responses with clear justifications to convince user.

\item \textbf{Personalized Requirement} evaluates the system's capability to incorporate individual user preferences and historical patterns, ensuring recommendations align with user habits and preferences.

\item \textbf{Without Factual Hallucination} serves as a critical safety check, identifying instances where the AI system generates non-existent services or logically inconsistent outputs that could mislead users.
\end{itemize}

\subsection{Case Study}
Due to ethic requirement, we cannot release the data used for training.
To ease reader understanding, we provide a real-world user query, and corresponding agent's response here.

Specifically, we select several real-world user case from the deployed travel planning service on DiDi Enterprise Solution App.
The illustrative case hopes to facilitate the reader's understanding of how DeepTravel enable TP agent autonomously plan, execute tool, and reflect on tool response for exploring and refine its intermediate steps throughout multi-turn reasoning process.

As illustrated in Table \ref{case_study}, the TP agent—powered by DeepTravel-32B—follows a multi-turn tool-integrated reasoning. Concretely: (1) Intent understanding and constraints grounding.
(2) Tool planning: Before calling tools, the agent drafts a tool plan with time buffers and fallback branches. (3) Tool execution with schema-aligned calls: All tool calls are structured with explicit parameters and expected fields, ensuring reproducibility and traceability. (4) Tool-response interpretation and evidence binding: The agent parses returned fields and binds them to the generated itinerary via card-style evidence, avoiding hallucinations. (5) Reflection: If constraints are unmet (e.g., arrival risk due to transfer), the agent revises either the departure time window or the station choice, and re-calls tools with updated parameters.
(6) Finalize the travel Itinerary: The agent assembles a coherent plan with justification, alternative options, and risk notes, each item linked to tool evidence.

\begin{figure*}[t]
\begin{AIbox}{Case Study: Reasoning Process of DeepTravel-32B}
\label{case_study}
\textbf{User Query}: Travel from Suzhou to the Wuhan Conference Center on June 26th and need to arrive before 3:00 PM. I'd like to stay near the riverside area. Please design a travel plan..
\\

\textbf{Agentic Travel Planning}:
\begin{verbatim}<think>Ok. So, I first need to confirm the user's needs. The user plans to travel from Suzhou to Wuhan Conference
...
We also generate the corresponding card information to ensure that the data comes from the tool's return results.
</think>
<tool_call_thinking>Search for direct high-speed trains from Suzhou to Wuhan, departing before 12:00 PM 
...
and allowing 1 hour for city transportation)
</tool_call_thinking>
<tool_call>train_search(depart_station="Suzhou", arrive_station="Wuhan", depart_date="2025-06-26")
</tool_call>
<tool_response>...</tool_response>
<tool_response_thinking>Several high-speed trains were found, but
...
</tool_response_thinking>
<tool_call_thinking>Search for hotels near Wuhan River Beach check-in date is June 26, check-out date is June 27
...
</tool_call_thinking>
<tool_call>hotel_search(city_name="Wuhan", checkin_date="2025-06-26", checkout_date="2025-06-27")
</tool_call>
<tool_response>...</tool_response>
<tool_response_thinking>I found several hotels near the river
...
</tool_response_thinking>
<answer>
### Outbound Recommendations (June 26, 2025)
#### G3124 High-Speed Train (Suzhou → Wuhan)
- **Recommendation**: Departs at 06:54, arrives at Wuhan Station
...

#### D3056 High-Speed Train (Suzhou → Hankou)
- **Recommendation**: Departs at 08:06, arrives at Hankou Station
...

### Hotel Recommendations
#### Qingchen Yuejiang Hotel (Jianghan Road Pedestrian Street)
- **Recommendation**: Only 233 meters from the river beach,
...

#### Le Bo 1946 Riverside Hotel (Hankou Jiangtan, Lihuangpih)
- **Recommendation**: River view rooms available, smart features
...

**Friendly Tips**:
1. Please book your shuttle in advance through the DiDi ES...
2. Please confirm room availability and breakfast policies...
</answer>

\end{verbatim}

\end{AIbox}
\end{figure*}

\end{document}